\documentclass{article}

\PassOptionsToPackage{round}{natbib}

\usepackage[final]{nips_2016}

\usepackage[utf8]{inputenc} 
\usepackage[T1]{fontenc}    
\usepackage{url}            
\usepackage{booktabs}       
\usepackage{amsfonts}       
\usepackage{nicefrac}       
\usepackage{microtype}      
\usepackage{graphicx}
\usepackage{color}

\title{Predicting Patient State-of-Health using Sliding Window and Recurrent Classifiers}

\author{
  Adam McCarthy\thanks{work completed whilst at University of Edinburgh}\\
  Department of Computer Science\\
  University of Oxford\\
  \texttt{adam.mccarthy@cs.ox.ac.uk}
  \And
  Christopher K.I. Williams\\
  School of Informatics, University of Edinburgh\\
  and Alan Turing Institute, London\\
  \texttt{ckiw@inf.ed.ac.uk}
}

\begin{document}

\maketitle

\begin{abstract}
Bedside monitors in Intensive Care Units (ICUs) frequently sound incorrectly, slowing response times and desensitising nurses to alarms \citep{Chambrin2001}, causing true alarms to be missed \citep{Hug2011}. We compare sliding window predictors with recurrent predictors to classify patient state-of-health from ICU multivariate time series; we report slightly improved performance for the RNN for three out of four targets.
\end{abstract}

\section{Introduction}

Intensive Care Units (ICUs) hold patients who need constant critical care, so it is essential that doctors are notified quickly of any change in the state of the patient through bedside monitoring. \cite{Imhoff2009} found that only 23\% of bedside alarms in ICU are clinically relevant, representing minimal improvement from a much earlier study \citep{Lawless1994}.

\cite{Georgatzis2015}, henceforth G\&W (2015), proposed a model which accurately classifies a number of clinical interventions which obscure patient vital signs. They discriminatively model these interventions using a Random Forest which takes sliding windows over the vital signs as input. This is a classification task for multiple binary targets, with a classifier for each modelled factor. They incorporate this classification into a Linear Dynamical System to infer the unobscured vital signs of the patient by maintaining a continuous state. 

We compare sliding window predictors such as Random Forests or Multilayer Perceptrons (MLPs) with recurrent predictors such as Recurrent Neural Networks (RNNs). The aim is to not only compare how they perform, but also to understand how they approach the problem by investigating the solutions that they have learned.
Sliding window predictors and recurrent predictors work in fundamentally different ways. Sliding window predictors make an independence assumption between each window, whereas recurrent predictors maintain dependence between the current timestep and all previous timesteps, prioritising more recent input. Gated RNN cells such as Long Short-Term Memory (LSTMs) \citep{Hochreiter1997} and Gated Recurrent Unit (GRUs) \citep{Chung2014} relax this independence assumption by learning to forget. They are effectively taking a sliding window over the data which they learn to adapt in length and structure in response to the input they receive. In this way, they may be able to capture longer-term dependencies than a sliding window predictor. We will show that the relaxation of the independence assumption is beneficial for the classification of ICU time series.

The sequence lengths for the recurrent predictor are unusually large at up to 153,678 timesteps. Related work from \cite{Lipton2015} uses sequences of up to around length 5,000 and \cite{Che2016} use sequences of up to length 150. In the broader RNN literature, \cite{Pouget-Abadie2014} report issues with sequences of over length 100 for neural machine translation and \cite{Le2015} report difficulty with sequences of length greater than 300. They rely on capturing the entirety of the dynamics of the input sequence in the hidden state, which limits the length of the sequences that can be processed. Additionally, gated RNN cells mitigate the \emph{vanishing gradient problem} \citep{Bengio1994} by learning to forget. We take a different approach, more akin to \cite{Zaremba2015}, where we output the classification at each timestep.

To the best of our knowledge this is the first example of using RNNs to classify ICU multivariate time series at every timestep. Other related work \citep[e.g.][]{Lipton2015, Che2016} generates a single label to represent the entire sequence once the whole input sequenced has been processed.

\section{Experiments}

\paragraph{Data} The dataset for these experiments is as used by \cite{LalParthaWilliamsChristopherK.I.GeorgatzisKonstantinosHawthorneChristopherMcMonaglePaulPiperIanShaw2015}. It incorporates twenty-seven adults admitted to the Neuro ICU at the Southern General Hospital in Glasgow. Information from four different channels was collected: electrocardiogram (ECG), systolic and diastolic arterial blood pressure (ABP), and systolic intracranial pressure (ICP). \cite{LalParthaWilliamsChristopherK.I.GeorgatzisKonstantinosHawthorneChristopherMcMonaglePaulPiperIanShaw2015} 
downsampled the readings to 1Hz, as this was shown by \cite{QuinnJohnA.2009} to be sufficient for ICU time series classification. The explicitly modelled factors are stable periods (lack of annotations), blood samples (BS), endotracheal suctioning (SC), and damped traces (DT). All other annotations were grouped into an `X factor'. The dataset is discussed in more detail by \cite{LalParthaWilliamsChristopherK.I.GeorgatzisKonstantinosHawthorneChristopherMcMonaglePaulPiperIanShaw2015}. 

\paragraph{Sliding window predictor} For the sliding window predictor, we trained an MLP. We followed G\&W (2015) by extracting a number of clinically-inspired features: least squared fits of a line of segments of the traces, an exponentially weighted moving average, the difference between the systolic and diastolic blood pressure channels (\emph{pulse pressure}), and the first order differences of the sequences.The pulse pressure is important for detecting the damped trace event. We therefore normalised the blood pressure channels together by taking the mean and standard deviations for both channels combined. We then subtracted this overall mean from both channels and divided both channels by the overall standard deviation. This meant that systolic ABP was generally positive and diastolic ABP was generally negative. The heart rate (HR) and ICP channels were normalised individually using the same process. We supplied only the ABP channels to the blood sample and damped trace predictors, and all channels in other cases.

We included a layer with linear activation functions and initially set the weights to extract the G\&W features (see \cite{McCarthy2016} for weights after backpropagation). To operate on segments of the input sequence, connections were removed as necessary. This fed into a number of Rectified Linear Unit (ReLU) layers and finally, a sigmoid output layer trained to output $P(f_t|x_{t-l:t+r})$, where $f_t$ is the modelled factor at time $t$ and $x_{t-l:t+r}$ is a sliding window over the input, where $l$ is the amount of past context and $r$ is the amount of future context.

\paragraph{Recurrent predictor}
\label{sec:exp-recurrent}

The recurrent predictor was composed of GRU hidden cells connected to a single sigmoid output unit. We used GRU cells instead of LSTMs as their simpler structure made them easier to analyse. \cite{Chung2014} demonstrated their performance was comparable over a range of problems. We trained using Truncated Backpropagation-Through-Time \citep{Graves2013}, with a length of 256 timesteps to make computation tractable. To investigate whether a recurrent predictor could learn long term correlations in the input, we directly supplied the input channels instead of extracting features. The Diastolic ABP was supplied to both the blood pressure and damped trace factors. For the blood sample factor we supplied the Systolic ABP and for the damped trace factor we supplied the pulse pressure. We did not supply the Systolic ABP to the damped trace factor in order to avoid multicollinearity. For the remaining factors, we supplied all channels.

We found with the MLP that including future context, $r$, of up to 10 seconds improved classification, in keeping with G\&W (2015). We therefore delayed the targets by 10 seconds during training to mirror this. Because gated RNN cells learn to forget, they are effectively learning a sliding window, but unlike sliding window predictors they adapt the size and structure of the sliding window based on the input they receive. Hyperparameter selection can learn an optimal sliding window length for a sliding window predictor, but it cannot adapt the sliding window structure to the extent that a gated RNN cell can.

\paragraph{Methodology}
\label{sec:methodology}

We trained both models using nested cross-validation \citep{Varma2006} to compare their overall performance. In our case, this involves performing 3-fold cross-validation in an outer loop and then performing leave-one-patient-out cross-validation on each fold as an inner loop. We perform hyperparameter selection in this inner loop and then assess performance with the optimal hyperparameters on the test set from the outer loop. This ensures that hyperparameter selection and performance assessment occurs on different data. \cite{Varma2006} found that cross-validation optimistically biases reported accuracy, whereas nested cross-validation reduces this bias. We evaluate the models by plotting Receiver Operating Characteristic curves on the concatenation of predictions in the three test sets and then compute the area under the curve.

We performed Early Stopping in all of the experiments. Every five epochs we calculated the cost on the validation set and compared with the cost five epochs prior. If the validation cost worsened on two successive occasions, we restored the weights which produced the minimum validation cost and terminated training. We did not use other regularisation techniques such as Dropout \citep{Srivastava2014} because applications of Dropout to RNNs \citep[e.g][]{Zaremba2015, Krueger2016} are relatively novel and less well understood. We used the Adam optimisation routine of \cite{Kingma2014} in all cases.

The input sequences were sparse, with only a small proportion of timesteps containing annotations. This class imbalance would make training difficult, so we used the annotated events as input sequences. However, we ensured that events from the same patient were never in the training and test set. These produced an even distribution of classes, including stable periods. We included context information of length equal to the event on either side, under the assumption that this would contain stable periods.

\paragraph{Hyperparameter selection}

These predictors have a number of hyperparameters which need to be set in order to create the optimal model. For the sliding window predictor, these were the number of hidden layers in $\{1,2,3\}$, the number of hidden units $h$ ($4 \leq h \leq 2048$), the length of the segments $4 \leq l \leq 49$ and $0 \leq r \leq 10$, and the learning rate $\mu$ ($0.001 \leq \mu \leq 0.1$). The number of hidden units was equal for all hidden layers. For the recurrent predictor, we optimised the number of hidden cells $c$ in each layer ($8 \leq c \leq 128$) and the learning rate ($0.001 \leq \mu \leq 0.1$). We include optimal hyperparameters in Appendix~\ref{app:optimal}.

G\&W (2015) performed a grid search over hyperparameters, but neural networks contain many more hyperparameters than random forests. Instead, we used Bayesian optimisation by fitting a Gaussian Process prior over our observations and generating new hyperparameters using an acquisition function which computes the expected improvement \citep{Snoek2012, snoek2014input, Gelbart2014}. This improved the selection time considerably.

\section{Results}

\begin{figure}[t]
    \centering
    \includegraphics[width=0.245\textwidth]{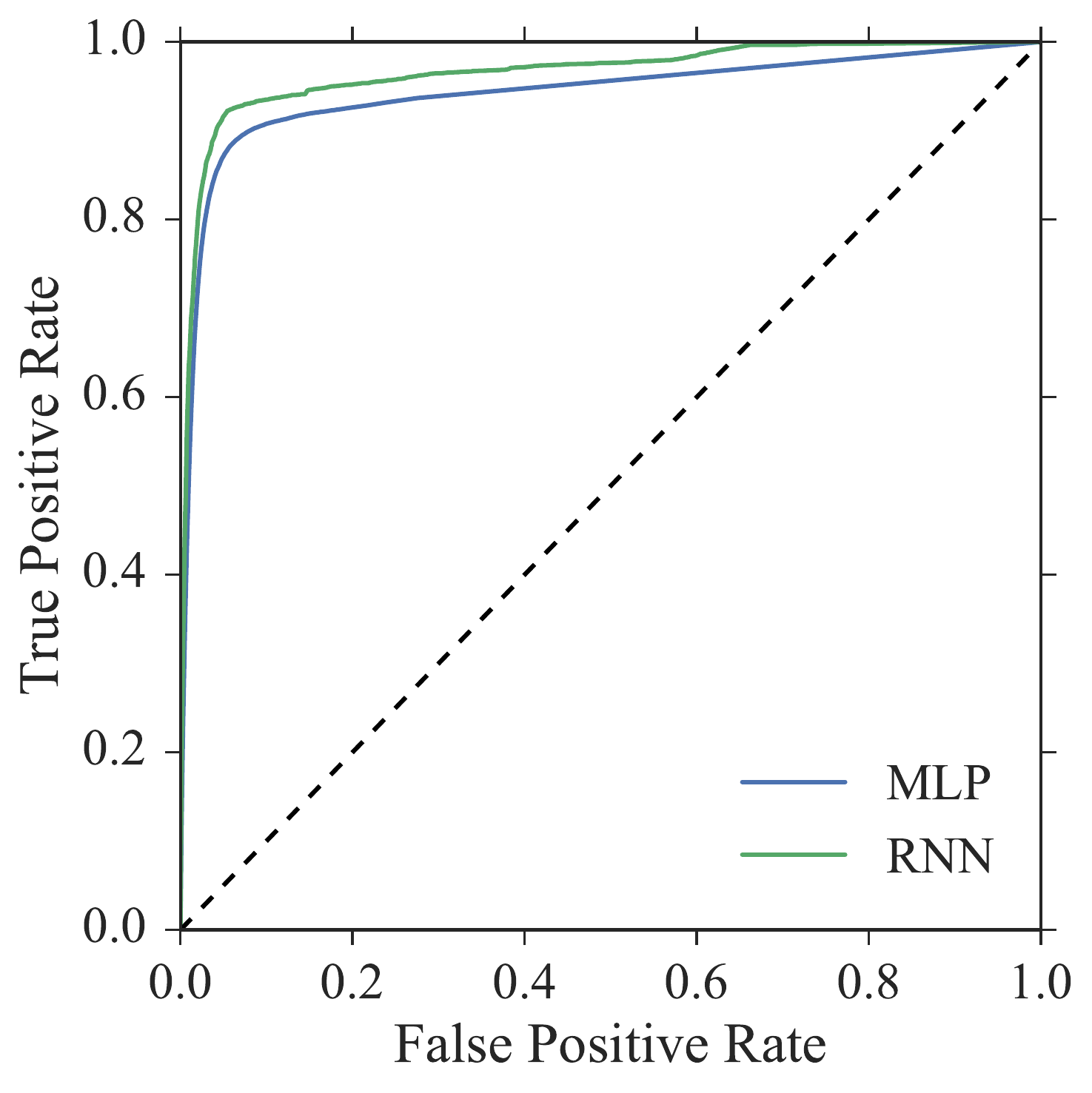}
    \includegraphics[width=0.245\textwidth]{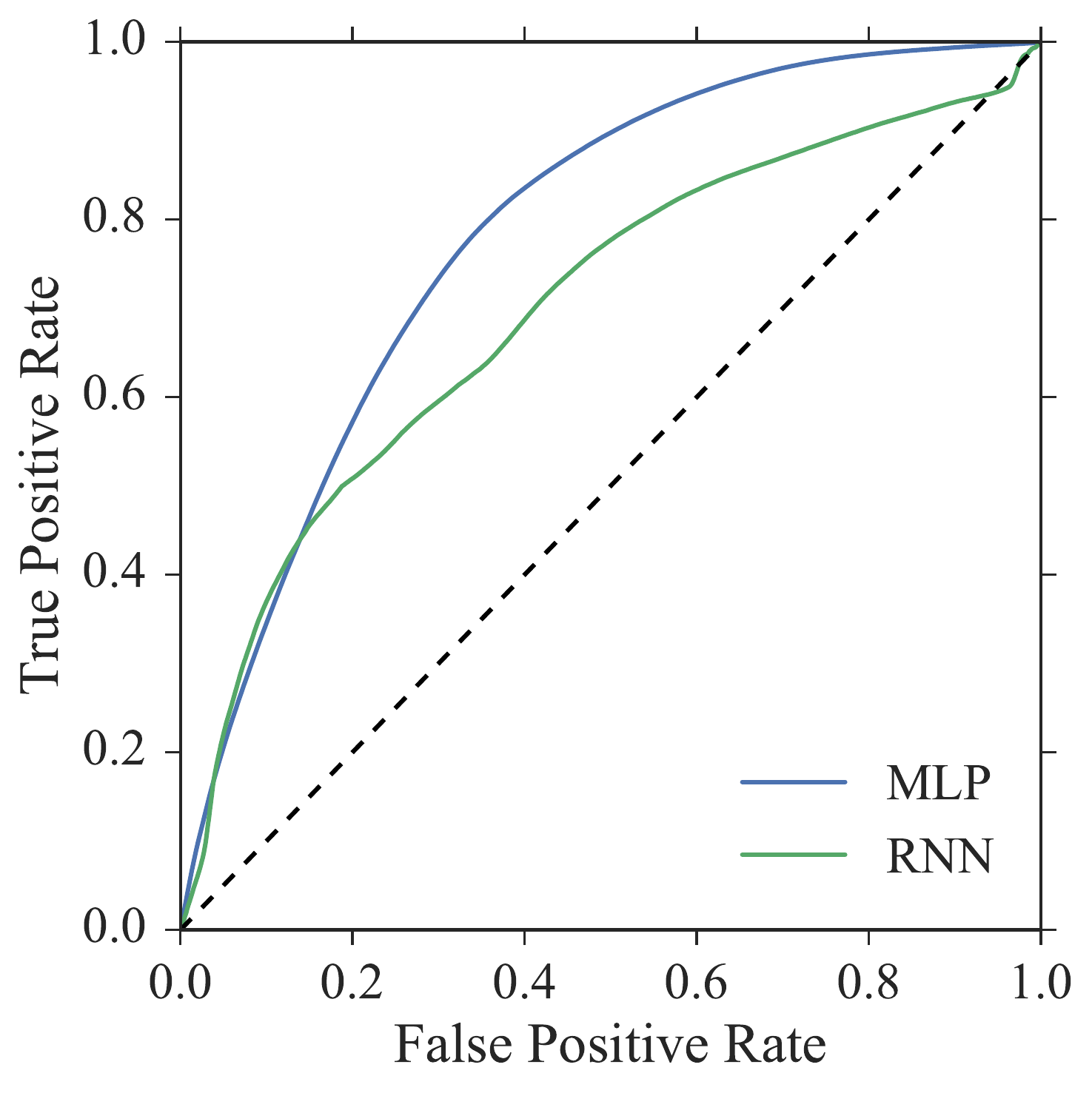}
    \includegraphics[width=0.245\textwidth]{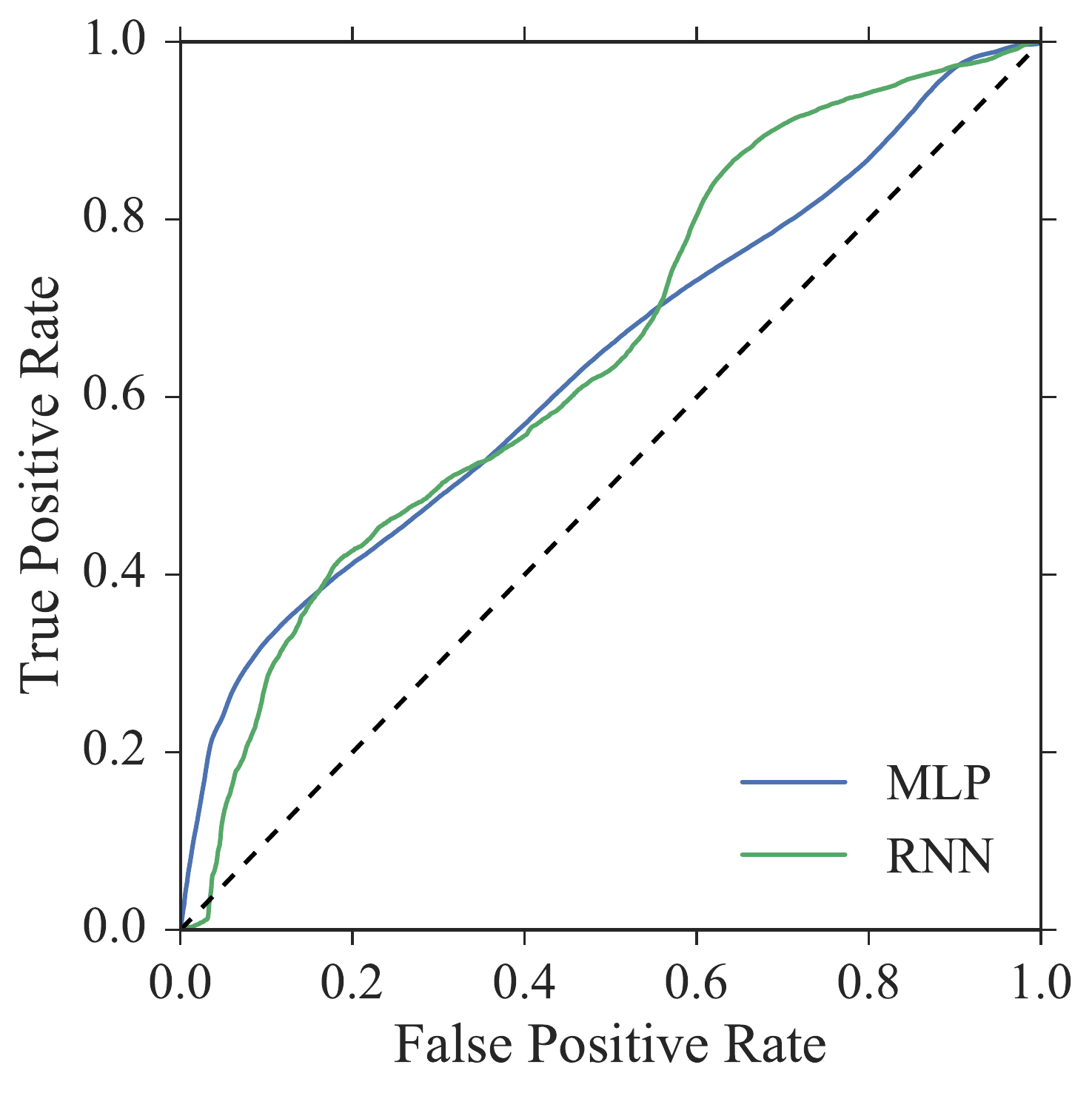}
    \includegraphics[width=0.245\textwidth]{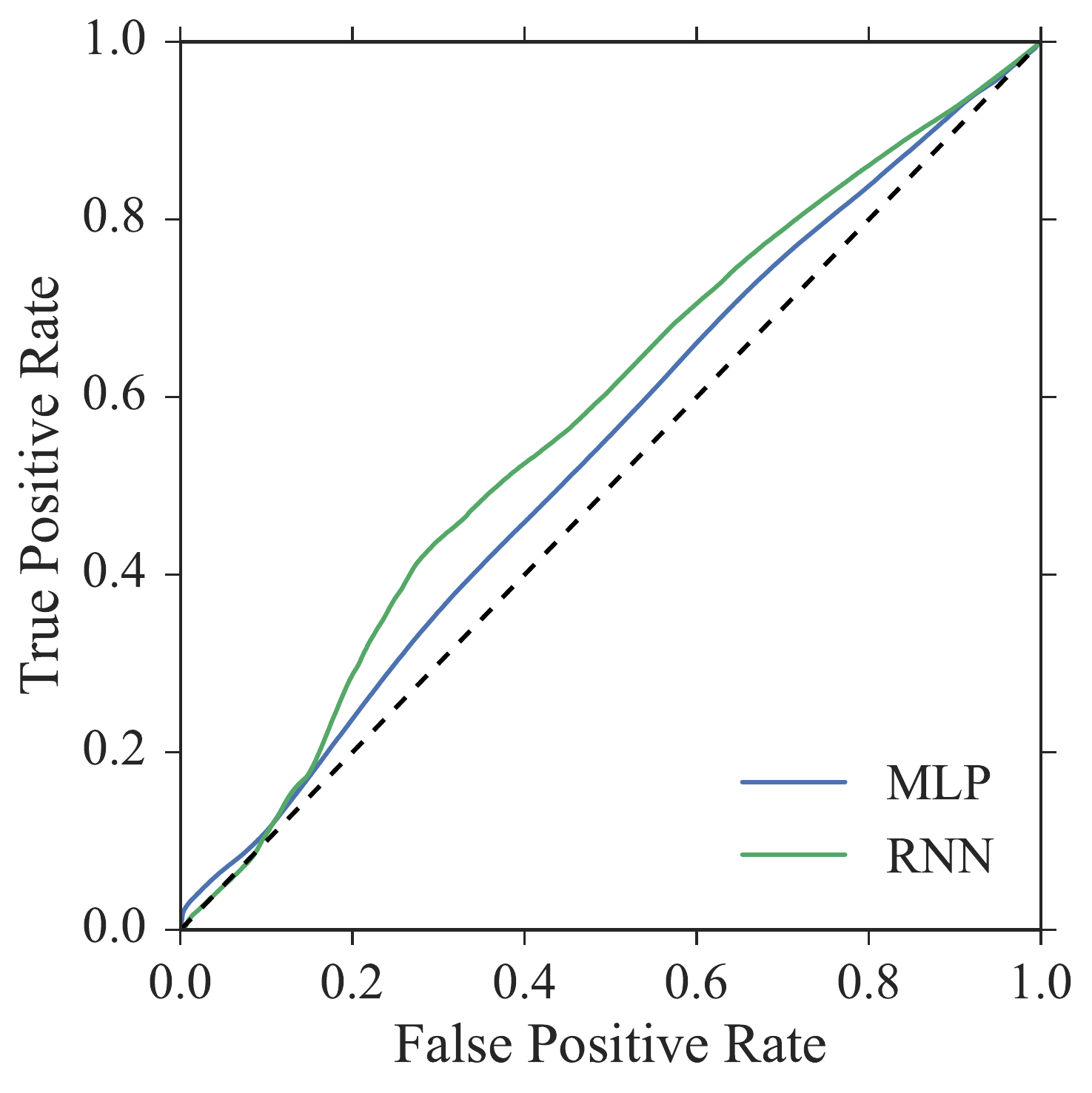}

    \begin{minipage}{0.245\textwidth}
        \centering
        a) Blood Sample
    \end{minipage}
    \begin{minipage}{0.245\textwidth}
        \centering
        b) Damped Trace
    \end{minipage}  
    \begin{minipage}{0.245\textwidth}
        \centering
        c) Suction Event
    \end{minipage}
    \begin{minipage}{0.245\textwidth}
        \centering
        d) X
    \end{minipage}   
    
    \caption{ROC curves on the Adult ICU Dataset from \cite{LalParthaWilliamsChristopherK.I.GeorgatzisKonstantinosHawthorneChristopherMcMonaglePaulPiperIanShaw2015}. }
    \label{fig:roc}
\end{figure}

In this section we will first report overall performance and then investigate the solutions learned by the models in order to interpret how they have learned.
\begin{table}[ht]
\caption{Comparison of DSLDS, MLP and RNN performance for the Adult ICU dataset.}
\label{tab:overall}
\begin{center}
\begin{tabular}{c c l l l l}
\multicolumn{1}{c}{\bf AUC}  \hspace{2mm}  &\multicolumn{1}{c}{BS} &\multicolumn{1}{c}{DT} &\multicolumn{1}{c}{SC} &\multicolumn{1}{c}{X}
\\ \hline \\
DSLDS \citep{LalParthaWilliamsChristopherK.I.GeorgatzisKonstantinosHawthorneChristopherMcMonaglePaulPiperIanShaw2015}\hspace{2mm} & $0.94$ & $\mathbf{0.78}$ & $0.64$ & $0.56$\\
MLP \hspace{2mm} & $0.94$ & $\mathbf{0.78}$ & $0.63$ & $0.54$\\
RNN \hspace{2mm} & $\mathbf{0.97}$ & $0.71$ & $\mathbf{0.65}$ & $\mathbf{0.58}$
\end{tabular}
\end{center}
\end{table}
\vspace{-1em}
\paragraph{Performance} AUC scores for the MLP and RNN on the Adult ICU Dataset from \cite{LalParthaWilliamsChristopherK.I.GeorgatzisKonstantinosHawthorneChristopherMcMonaglePaulPiperIanShaw2015}  are shown in Table~\ref{tab:overall} and ROC curves are included in Figure~\ref{fig:roc}. The MLP and the random forest used by \cite{LalParthaWilliamsChristopherK.I.GeorgatzisKonstantinosHawthorneChristopherMcMonaglePaulPiperIanShaw2015} show very similar performance. The RNN achieved slightly improved results for the Blood Sample, Suction Event and X Factors.

The RNN could not match the performance of the MLP on damped traces and we have a number of hypotheses to explain this. The first is that the largest RNN we could run on one GPU had 128 hidden cells, whereas the largest MLP has 2048 hidden units. Whilst this is fine for temporal correlations, as the RNN hidden states are maintained through each timestep which gives it significantly more parameters, only the candidate activation can capture the pulse pressure at each timestep. Additionally, much of this information is lost when it passes through the update gate. Another possibility is that the damped trace sequences, at 153,678 timesteps long (42 hours), are too long for the RNN to fully capture their dynamics. They may vary through the period due to factors such as medical interventions or circadian rhythm. In comparison, the longest sequences for the Blood Sample, Endotracheal Suction and X factor sequences were 1,149 timesteps (19 mins), 13,176 timesteps (3.65 hrs), and 49,740 timesteps (13.81 hrs) respectively. Finally, we postulate in Section~\ref{sec:discussion} that the RNNs are better at adapting to the baseline physiology of the patient. This can cause them to not classify some timesteps as damped traces, even if the annotations indicate that that they are.

\section{Discussion}
\label{sec:discussion}

To investigate how the predictors learned, we plotted the predictions, saliency maps between the input and the output \citep{Morch1995}, and saliency maps between the activations \citep{Erhan2009}. Empirically, we made the following observations, which could warrant further study (see \cite{McCarthy2016} sections 4 and 5 for further details).

\paragraph{Adapting to baseline physiology} Sliding window predictors incorrectly classify low pulse pressures as damped traces, whereas recurrent predictors wait for a reduction in pulse pressure before making a classification. This is important in ICU due to high vital sign variability between patients.

\paragraph{Classifying long-term events}

Events are often obscured by other pathology. RNNs can maintain state through these disturbances, whereas sliding window predictors will often only classify the beginning and end of long events. However, this state maintenance can cause RNN hidden states to decay too slowly when events finish if they are not well delineated.

\paragraph{Noisy input sequences}

The RNN was better at handling very volatile input because the hidden state causes the predictions to effectively be smoothed, in comparison to the MLP which produces very volatile predictions in response to this volatility.

\paragraph{Computational complexity}

RNNs are much more computationally expensive to train. It is necessary to zero pad the sequences to the length of the longest sequence if one wishes to use a BLAS library with a batch size greater than one. Time spent processing zeros is wasted and their activations consume GPU RAM, limiting the size of models which can be trained. Solutions to this include using a smaller batch size, sacrificing parallelism, or multiplying sparse matrices which is slow. Sequences could also be segmented, improving parallelism at the cost of not capturing long-term dependencies.

\section{Conclusion}

In this report we have trained MLP and RNNs to compare sliding window and recurrent predictors in the classification of ICU time series. G\&W (2015) built an ensemble of two sliding window predictors: the FSLDS and DSLDS, which improved performance compared to the individual models. Adding a recurrent predictor to this mixture may improve these results further.

We hope that the insight into how the models learn to approach the problem gained in this report will help the construction of better bedside alarms for the ICU in the future. Ideally any follow up to \cite{Lawless1994} and \cite{Imhoff2009} will find that the wolves are no longer crying.

\newpage
Acknowledgements: The work of CW is supported in part by EPSRC grant
EP/N510129/1.
\bibliographystyle{abbrvnat}
{\small \bibliography{main}}

\begin{thebibliography}{26}
\providecommand{\natexlab}[1]{#1}
\providecommand{\url}[1]{\texttt{#1}}
\expandafter\ifx\csname urlstyle\endcsname\relax
  \providecommand{\doi}[1]{doi: #1}\else
  \providecommand{\doi}{doi: \begingroup \urlstyle{rm}\Url}\fi

\bibitem[Bengio et~al.(1994)Bengio, Simard, and Frasconi]{Bengio1994}
Y.~Bengio, P.~Simard, and P.~Frasconi.
\newblock {Learning Long-Term Dependencies with Gradient Descent is Difficult}.
\newblock \emph{IEEE Transactions on Neural Networks}, 5\penalty0 (2):\penalty0
  157--166, 1994.
\newblock ISSN 19410093.
\newblock \doi{10.1109/72.279181}.

\bibitem[Chambrin(2001)]{Chambrin2001}
M.~C. Chambrin.
\newblock {Alarms in the intensive care unit: how can the number of false
  alarms be reduced?}
\newblock \emph{Critical Care}, 5\penalty0 (4):\penalty0 184--188, 2001.
\newblock ISSN 1364-8535.
\newblock \doi{10.1186/cc1021}.

\bibitem[Che et~al.(2016)Che, Purushotham, Cho, Sontag, and Liu]{Che2016}
Z.~Che, S.~Purushotham, K.~Cho, D.~Sontag, and Y.~Liu.
\newblock {Recurrent Neural Networks for Multivariate Time Series with Missing
  Values}.
\newblock \emph{arXiv:1606.01865}, 2016.

\bibitem[Chung et~al.(2014)Chung, Gulcehre, Cho, and Bengio]{Chung2014}
J.~Chung, C.~Gulcehre, K.~Cho, and Y.~Bengio.
\newblock {Empirical Evaluation of Gated Recurrent Neural Networks on Sequence
  Modeling}.
\newblock \emph{arXiv:1412.3555v1}, pages 1--9, 2014.

\bibitem[Erhan et~al.(2009)Erhan, Courville, Bengio, and Vincent]{Erhan2009}
D.~Erhan, A.~Courville, Y.~Bengio, and P.~Vincent.
\newblock {Visualizing Higher Layer Features of a Deep Network}.
\newblock \emph{Technical Report 1341, DIRO, Universite de Montreal}, 2009.

\bibitem[Gelbart et~al.(2014)Gelbart, Snoek, and Adams]{Gelbart2014}
M.~A. Gelbart, J.~Snoek, and R.~P. Adams.
\newblock {Bayesian Optimization with Unknown Constraints}.
\newblock \emph{arXiv:1403.5607v1}, pages 1--14, 2014.
\newblock ISSN 1098-6596.
\newblock \doi{10.1017/CBO9781107415324.004}.

\bibitem[Georgatzis and Williams(2015)]{Georgatzis2015}
K.~Georgatzis and C.~K.~I. Williams.
\newblock {Discriminative Switching Linear Dynamical Systems applied to
  Physiological Condition Monitoring}.
\newblock \emph{Proceedings of Uncertainty in AI}, 2015.

\bibitem[Graves(2013)]{Graves2013}
A.~Graves.
\newblock {Generating sequences with recurrent neural networks}.
\newblock \emph{arXiv:1308.0850}, pages 1--43, 2013.
\newblock ISSN 18792782.
\newblock \doi{10.1145/2661829.2661935}.

\bibitem[Hochreiter and Schmidhuber(1997)]{Hochreiter1997}
S.~Hochreiter and J.~Schmidhuber.
\newblock {Long Short-Term Memory}.
\newblock \emph{Neural Computation}, 9\penalty0 (8):\penalty0 1735--1780, 1997.
\newblock ISSN 0899-7667.
\newblock \doi{10.1162/neco.1997.9.8.1735}.

\bibitem[Hug et~al.(2011)Hug, Clifford, and Reisner]{Hug2011}
C.~W. Hug, G.~D. Clifford, and A.~T. Reisner.
\newblock {Clinician blood pressure documentation of stable intensive care
  patients: an intelligent archiving agent has a higher association with future
  hypotension.}
\newblock \emph{Critical Care Medicine}, 39\penalty0 (5):\penalty0 1006--14,
  2011.
\newblock \doi{10.1097/CCM.0b013e31820eab8e}.

\bibitem[Imhoff and Fried(2009)]{Imhoff2009}
M.~Imhoff and R.~Fried.
\newblock {The crying wolf: Still crying?}
\newblock \emph{Anesthesia and Analgesia}, 108\penalty0 (5):\penalty0
  1382--1383, 2009.
\newblock ISSN 00032999.
\newblock \doi{10.1213/ane.0b013e31819ed484}.

\bibitem[Kingma and Ba(2015)]{Kingma2014}
D.~Kingma and J.~Ba.
\newblock {Adam: A Method for Stochastic Optimization}.
\newblock In \emph{International Conference on Learning Representations}, pages
  1--13, 2015.

\bibitem[Krueger and Memisevic(2016)]{Krueger2016}
D.~Krueger and R.~Memisevic.
\newblock {Regularizing RNNs by Stabilizing Activations}.
\newblock \emph{ICLR}, pages 1--8, 2016.

\bibitem[Lal et~al.(2015)Lal, Williams, Konstantinos, Hawthorne, McMonagle,
  Piper, and
  Shaw]{LalParthaWilliamsChristopherK.I.GeorgatzisKonstantinosHawthorneChristopherMcMonaglePaulPiperIanShaw2015}
P.~Lal, C.~K.~I. Williams, G.~Konstantinos, P.~Hawthorne, C.~McMonagle,
  I.~Piper, and M.~Shaw.
\newblock {Detecting Artifactual Events in Vital Signs Monitoring Data}.
\newblock Technical report, University of Edinburgh and University of Glasgow,
  2015.

\bibitem[Lawless(1994)]{Lawless1994}
S.~T. Lawless.
\newblock {Crying wolf: false alarms in a pediatric intensive care unit.},
  1994.
\newblock ISSN 0090-3493.

\bibitem[Le et~al.(2015)Le, Jaitly, and Hinton]{Le2015}
Q.~V. Le, N.~Jaitly, and G.~E. Hinton.
\newblock {A Simple Way to Initialize Recurrent Networks of Rectified Linear
  Units}.
\newblock \emph{arXiv:1504.00941}, 2015.

\bibitem[Lipton et~al.(2015)Lipton, Kale, Elkan, and Wetzell]{Lipton2015}
Z.~C. Lipton, D.~C. Kale, C.~Elkan, and R.~Wetzell.
\newblock {Learning to Diagnose with LSTM Recurrent Neural Networks}.
\newblock In \emph{ICLR}, pages 1--18, 2015.

\bibitem[McCarthy(2016)]{McCarthy2016}
A.~McCarthy.
\newblock {An Evaluation of Sliding Window and Recurrent Predictors for the
  Classification of ICU Time Series. MSc dissertation. University of
  Edinburgh.}
\newblock 2016.

\bibitem[M{\o}rch et~al.(1995)M{\o}rch, Kjems, Hansen, Svarer, Law, Lautrup,
  Strother, and Rehm]{Morch1995}
N.~J.~S. M{\o}rch, U.~Kjems, L.~K. Hansen, C.~Svarer, I.~Law, B.~Lautrup, S.~C.
  Strother, and K.~Rehm.
\newblock {Visualization of Neural Networks Using Saliency Maps}.
\newblock In \emph{Proceedings of 1995 IEEE International Conference on Neural
  Networks}, volume~4, pages 2085--2090, 1995.
\newblock ISBN 0-7803-2768-3.
\newblock \doi{10.1109/ICNN.1995.488997}.

\bibitem[Pouget-Abadie et~al.(2014)Pouget-Abadie, Bahdanau, van Merrienboer,
  Cho, and Bengio]{Pouget-Abadie2014}
J.~Pouget-Abadie, D.~Bahdanau, B.~van Merrienboer, K.~Cho, and Y.~Bengio.
\newblock {Overcoming the Curse of Sentence Length for Neural Machine
  Translation using Automatic Segmentation}.
\newblock \emph{arXiv:1409.1257}, 2014.

\bibitem[Quinn et~al.(2009)Quinn, Williams, and McIntosh]{QuinnJohnA.2009}
J.~A. Quinn, C.~K. Williams, and N.~McIntosh.
\newblock {Factorial Switching Linear Dynamical Systems Applied to
  Physiological Condition Monitoring}.
\newblock \emph{IEEE Transactions on Pattern Analysis and Machine
  Intelligence}, 31\penalty0 (9):\penalty0 1537--1551, 2009.
\newblock ISSN 0162-8828.
\newblock \doi{10.1109/TPAMI.2008.191}.

\bibitem[Snoek et~al.(2012)Snoek, Larochelle, and Adams]{Snoek2012}
J.~Snoek, H.~Larochelle, and R.~P. Adams.
\newblock {Practical Bayesian Optimization of Machine Learning Algorithms}.
\newblock In F.~Pereira, C.~J.~C. Burges, L.~Bottou, and K.~Q. Weinberger,
  editors, \emph{Advances in Neural Information Processing Systems 25}, pages
  2951--2959. Curran Associates, Inc., 2012.

\bibitem[Snoek et~al.(2014)Snoek, Swersky, Zemel, and Adams]{snoek2014input}
J.~Snoek, K.~Swersky, R.~S. Zemel, and R.~P. Adams.
\newblock {Input Warping for Bayesian Optimization of Non-Stationary
  Functions.}
\newblock In \emph{ICML}, pages 1674--1682, 2014.

\bibitem[Srivastava et~al.(2014)Srivastava, Hinton, Krizhevsky, Sutskever, and
  Salakhutdinov]{Srivastava2014}
N.~Srivastava, G.~E. Hinton, A.~Krizhevsky, I.~Sutskever, and R.~Salakhutdinov.
\newblock {Dropout : A Simple Way to Prevent Neural Networks from Overfitting}.
\newblock \emph{Journal of Machine Learning Research (JMLR)}, 15:\penalty0
  1929--1958, 2014.
\newblock ISSN 15337928.
\newblock \doi{10.1214/12-AOS1000}.

\bibitem[Varma and Simon(2006)]{Varma2006}
S.~Varma and R.~Simon.
\newblock {Bias in Error Estimation when using Cross-Validation for Model
  Selection.}
\newblock \emph{BMC Bioinformatics}, 7:\penalty0 91, 2006.
\newblock ISSN 1471-2105.
\newblock \doi{10.1186/1471-2105-7-91}.

\bibitem[Zaremba et~al.(2015)Zaremba, Sutskever, and Vinyals]{Zaremba2015}
W.~Zaremba, I.~Sutskever, and O.~Vinyals.
\newblock {Recurrent Neural Network Regularization}.
\newblock \emph{ICLR}, pages 1--8, 2015.

\end{thebibliography}

\newpage
\appendix
\section{Optimal hyperparameters}
\label{app:optimal}
\begin{table}[ht]
\caption{Optimal RNN hyperparameters. All networks had one hidden layer.}
\begin{center}
\begin{tabular}{c c l l}
\multicolumn{1}{c}{\bf Factor} &\multicolumn{1}{c}{\bf Outer Fold} \hspace{2mm} &\multicolumn{1}{c}{$c$} &\multicolumn{1}{c}{$\mu$}
\\ \hline \\
   & 1 & \hspace{2mm} $64$ & $0.001$\\
BS & 2 & \hspace{2mm} $15$ & $0.006625$\\
   & 3 & \hspace{2mm} $14$ & $0.009637$\\
\\ \hline \\
   & 1 & \hspace{2mm} $58$ & $0.003695$\\
DT & 2 & \hspace{2mm} $8$  & $0.001$\\
   & 3 & \hspace{2mm} $67$ & $0.001$\\
\\ \hline \\
   & 1 & \hspace{2mm} $8$  & $0.01$\\
SC & 2 & \hspace{2mm} $8$  & $0.01$\\
   & 3 & \hspace{2mm} $8$  & $0.01$\\
\\ \hline \\
   & 1 & \hspace{2mm} $64$ & $0.001$\\
X  & 2 & \hspace{2mm} $8$ & $0.001$\\
   & 3 & \hspace{2mm} $18$ & $0.003813$
\end{tabular}
\end{center}
\end{table}

\begin{table}[ht]
\caption{Optimal MLP hyperparameters. All networks had one hidden layer.}
\begin{center}
\begin{tabular}{c c l l l l}
\multicolumn{1}{c}{\bf Factor} &\multicolumn{1}{c}{\bf Outer Fold} \hspace{2mm} &\multicolumn{1}{c}{$h$} & \multicolumn{1}{c}{$l$} & \multicolumn{1}{c}{$r$} & \multicolumn{1}{c}{$\mu$}
\\ \hline \\
   & 1 & \hspace{2mm} $50$  & $9$ & $10$ & $0.0005$\\
BS & 2 & \hspace{2mm} $962$ & $9$ & $5$  & $0.007923$\\
   & 3 & \hspace{2mm} $763$ & $4$ & $0$  & $0.0005$\\
\\ \hline \\
   & 1 & \hspace{2mm} $246$ & $9$  & $10$ & $0.004294$\\
DT & 2 & \hspace{2mm} $882$ & $9$  & $0$  & $0.006438$\\
   & 3 & \hspace{2mm} $347$ & $14$ & $5$  & $0.007031$\\
\\ \hline \\
   & 1 & \hspace{2mm} $50$  & $9$ & $0$ & $0.0005$\\
SC & 2 & \hspace{2mm} $277$ & $4$ & $0$ & $0.00527$\\
   & 3 & \hspace{2mm} $64$  & $4$ & $0$ & $0.0005$\\
\\ \hline \\
   & 1 & \hspace{2mm} $50$  & $4$ & $0$ & $0.0005$\\
X  & 2 & \hspace{2mm} $525$ & $4$ & $0$ & $0.00525$\\
   & 3 & \hspace{2mm} $50$ & $4$ & $0$  & $0.0005$
\end{tabular}
\end{center}
\end{table}

\end{document}